\documentclass[aos,MSNbibl,citesort,dvips]{arximspdf}
\usepackage{algorithm}
\usepackage{algorithmic}
\usepackage{mathrsfs}
\usepackage{graphicx}

\doi{10.1214/12-AOS1009} 
\volume{40}
\issue{3}
\pubyear{2012}
\firstpage{1346}
\lastpage{1375}

\makeatletter

\newcommand{\norm}[1]{\Vert #1 \Vert}
\newcommand{\nbd}{\mathcal{N}}
\newcommand{\eqref}[1]{(\ref{#1})}
\newcommand{\Deg}{\operatorname{Deg}}

\newcommand{\Aug}{\operatorname{Aug}}
\newcommand{\watts}{\operatorname{Watts}}
\newcommand{\girth}{\operatorname{Girth}}
\newcommand{\LP}{\operatorname{LP}}
\newcommand{\ER}{\mathrm{ER}}
\newcommand{\reg}{\operatorname{Reg}}
 \newcommand{\indep}{\protect\mathpalette{\protect\independenT}{\perp}}
\def\independenT#1#2{\mathrel{\rlap{$#1#2$}\mkern2mu{#1#2}}}
\newcommand{\atanh}{\operatorname{atanh}}

\def\nn{\nonumber}

\newcommand{\Gmsc}{\mathcal{G}}

\def\hG{\widehat{G}}
\def\hP{\widehat{P}}
\def\hQ{\widehat{Q}}

\def\hnu{\widehat{\nu}}

\def\bfzero{\mathbf{0}}
\def\bfh{\mathbf{h}}
\def\bfJ{\mathbf{J}}
\def\bfX{\mathbf{X}}

\def\Cc{\mathcal{C}}
\def\Hc{\mathcal{H}}
\def\Qc{\mathcal{Q}}
\def\Sc{\mathcal{S}}
\def\Xc{\mathcal{X}}

\def\Nbb{\mathbb{N}}
\def\Rbb{\mathbb{R}}

\def\tilo{{\widetilde{o}}}
\def\tilomega{{\widetilde{\omega}}}

\newtheorem{theorem}{Theorem}

\newtheorem{lemma}{Lemma}

\newproclaim{remarks}{Remarks}

\newproclaim{definition}{Definition}

\newproclaim{remark}{Remark}
\newproclaim{example}{Example}

\def\nbd{\mathcal{N}}

\newcommand{\calX}{\mathcal{X}}
\newcommand{\calZ}{\mathcal{Z}}
\newcommand{\calY}{\mathcal{Y}}
\newcommand{\bX}{\mathbf{X}}
\newcommand{\calS}{\mathcal{S}}
\newcommand{\calN}{\mathcal{N}}
\newcommand{\Ind}{\mathbb{I}}

\newcommand{\poly}{\mathrm{poly}}
\newcommand{\threscondalgo}{\mathsf{CVDT}}
\newcommand{\threscmialgo}{\mathsf{CMIT}}

\makeatother

\begin{document}
\begin{frontmatter}

\title{High-dimensional structure estimation in Ising models: Local
separation criterion\thanksref{T1}}

\runtitle{Structure learning of Ising models}
\thankstext{T1}{An abridged version of this paper appears in Proc. of
NIPS 2011.}
\begin{aug}
\author[A]{\fnms{Animashree} \snm{Anandkumar}\corref{}\thanksref{t1}\ead[label=e1]{a.anandkumar@uci.edu}},
\author[B]{\fnms{Vincent Y. F.} \snm{Tan}\thanksref{t2,t3}\ead[label=e2]{tanyfv@i2r.a-star.edu.sg}},
\author[A]{\fnms{Furong}~\snm{Huang}\thanksref{t1}\ead[label=e3]{furongh@uci.edu}}
\and
\author[C]{\fnms{Alan S.} \snm{Willsky}\thanksref{t3}\ead[label=e4]{willsky@mit.edu}}

 \runauthor{Anandkumar, Tan, Huang and Willsky}

\thankstext{t1}{Supported by the setup funds at UCI and the AFOSR
Award FA9550-10-1-0310.}
\thankstext{t2}{Supported in part by A*STAR, Singapore.}
\thankstext{t3}{Supported in part by AFOSR under Grant FA9550-08-1-1080.}

\affiliation{University of California Irvine, Institute for Infocomm Research and National University
of Singapore, University of California Irvine and
Massachusetts Institute of Technology}
\address[A]{A. Anandkumar\\
F. Huang\\
Center for Pervasive Communications\\
\quad\& Computing\\
Electrical Engineering\\
\quad\& Computer Science Department\\
4408 Engineering Hall\\
Irvine, California 92697\\
USA\\
\printead{e1}\\
\phantom{E-mail: }\printead*{e3}}
\address[B]{V. Y. F. Tan\\
Institute for Infocomm Research\\
A*STAR Singapore\\
and\\
Department of Electrical\\
\quad and Computer Engineering\\
National University of Singapore\\
Singapore\\
\printead{e2}}

\address[C]{A. S. Willsky\\
Laboratory of Information\\
\quad\& Decision Systems\\
Stata Center, 77 Massachusetts Ave.\\
Cambridge, Massachusetts 02139\\
USA\\
\printead{e4}}

\end{aug}


%
\begin{abstract}
We consider the problem of high-dimensional Ising (graphical) model
selection. We propose a simple algorithm for structure estimation based
on the thresholding of the empirical conditional variation distances.
We introduce a novel criterion for tractable graph families, where this
method is efficient, based on the presence of sparse local separators
between node pairs in the underlying graph. For such graphs, the
proposed algorithm has a sample complexity of $n
=\Omega(J_{\min}^{-2} \log p)$, where $p$ is the number of variables,
and $J_{\min}$ is the minimum (absolute) edge potential in the model.
We also establish nonasymptotic necessary and sufficient conditions
for structure estimation.
\end{abstract}

%
\begin{keyword}[class=AMS]
\kwd[Primary ]{62H12}
\kwd[; secondary ]{05C80}.
\end{keyword}
\begin{keyword}
\kwd{Ising models}
\kwd{graphical model selection}
\kwd{local-separation property}.
\end{keyword}

\end{frontmatter}

\section{Introduction}

The use of probabilistic graphical models allows for succinct
representation of high-dimensional distributions, where the
conditional-independence relationships among the variables are
represented by a graph. Such models have found many applications in a
variety of areas, including computer vision~\cite{choidcvpr10},
bio-informatics~\cite{Durbin:book}, financial modeling
\cite{Chodetald0JMLR} and social networks \cite{grabowski2006ising}.
For instance, graphical models are employed for contextual object
recognition to improve detection performance based on object
co-occurrences~\cite{choidcvpr10} and for modeling opinion formation
and technology adoption in social networks
\cite{grabowski2006ising,laciana2010ising}.

A major challenge involving graphical models is structure estimation,
given samples drawn from the model. It is known that such a learning
task is NP-hard \cite{Kar01,Bogdanov&etal:Rand}. This challenge is
compounded in the {\em high-dimensional} regime, where the number of
available observations is typically much smaller than the number of
dimensions (or variables). It is thus imperative to design efficient
algorithms for structure estimation of graphical models with low
sample complexity.

In their seminal work, Chow and Liu presented an efficient algorithm
for structure estimation of tree-structured graphical models based on a
maximum weight spanning tree algorithm \cite{Chow&Liu:68IT}. Since
then, various algorithms have been proposed for structure estimation of
sparse graphical models. They can be broadly classified into two
categories: combinatorial algorithms
\cite{BreslerdetaldRand,SanghavidetadAllerton10} and those based on
convex relaxation
\cite
{Mei06,Ravikumardetad08Arxiv,Ravikumardetald08Stat,Chandrasekaran:10latent}.
The former approach is typically based on certain local tests on small
groups of data, and then combining them to output a graph structure,
while the latter approach involves solving a penalized convex
optimization problem. See Section~\ref{sec:related} for a detailed
discussion of these approaches.

\setcounter{footnote}{4}

In this paper, we propose a novel local algorithm and analyze its
performance for structure estimation of Ising models, which are
pairwise binary graphical models. Our proposed algorithm circumvents
one of the primary limitations of existing local algorithms
\cite{BreslerdetaldRand,SanghavidetadAllerton10} for consistent
estimation in high-dimensions---that the graphs have a bounded degree
as the number of nodes $p$ tends to infinity. We give a precise
characterization of the class of graphs which can be consistently
recovered by our algorithm with low computational and sample
complexities. We demonstrate that a fundamental property shared by
these graphs is that they have {\em sparse local vertex separators}
between any two nonneighbors in the graph. A wide variety of graphs
satisfy this property. These include large girth graphs, the
Erd\H{o}s--R\'{e}nyi random graphs\footnote{The Erd\H{o}s--R\'{e}nyi
graphs have sparse local vertex separators asymptotically almost surely
(a.a.s.) with respect to the random graph measure. Indeed, whenever we
mention ensembles of random graphs in the sequel, our statements are
taken to hold a.a.s. }~\cite{Bollobas:book} and the power-law
graphs~\cite{Chung:book}, as well as graphs with short cycles such as
the small-world graphs~\cite{watts1998collective} and other hybrid
graphs~\cite[Chapter 12]{Chung:book}.

Our results are applicable in the realms of social networks,
bio-informatics, computer vision and so on. Here, we elaborate on
its relevance to social networks. The aforementioned graphs (i.e.,
the power-law and the small-world graphs) have been employed
extensively for modeling the topologies of social networks
\cite{Newman:02,albert2002statistical}. More recently, Ising models on
such topologies have been employed for modeling various phenomena in
social networks \cite{vega2007complex}, such as opinion formation
\cite{galam1997rational,grabowski2006ising,Liu:10Allerton} and
technology adoption \cite{laciana2010ising}. A concrete example is the
use of an Ising model for the U.S. senate voting network \cite{senate}.
The nodes of the graph represent the senators, and the data are the
voting decisions made by the senators. Estimating the graph reveals
interesting relationships between the senators and the effect of
political affiliations on their decisions. Similarly, in many other
scenarios (e.g., online social networks), we have access to a sequence
of measurements at the nodes of the network. For instance, we may
gather the opinions of different users or measure the popularity of new
technologies. As a first-order approximation, we can regard such a
sequence of measurements as being independent and identically
distributed (i.i.d.) samples drawn from an Ising model. Our findings
imply that the topology of such social-network models can be
efficiently estimated under some mild and transparent conditions.


\subsection{Summary of results}

Our main contributions in this work are threefold. We propose a simple
algorithm for structure estimation of Ising models. The algorithm is
based on approximate conditional independence testing based on
conditional variation distances. Second, we derive sample complexity
results for consistent structure estimation in high dimensions. Third,
we prove novel lower bounds on the sample complexity required for any
learning algorithm to be consistent for model selection.

We propose an algorithm for structure estimation, termed as conditional
variation distance thresholding ($\threscondalgo$), which tests if two
nodes are neighbors by searching for a node set which (approximately)
separates them in the underlying Markov graph. It first computes the
minimum empirical conditional variation distance in
\eqref{eqn:empcondvariation} of a given node pair over conditioning
sets of bounded cardinality $\eta$. Second, if the minimum exceeds a
given threshold (depending on the number of samples $n$ and the number
of nodes $p$), the node pair is declared as an edge. This test has a
computational complexity of $O(p^{\eta+2})$. Thus, the computational
complexity is low if $\eta$ is small. Further, it requires only
low-order statistics (up to order $\eta+2$). We establish that the
parameter $\eta$ is a bound on the size of local vertex-separators
between any two nonneighbors in the graph, and is small for many
common graph families, introduced before.

We establish that under a set of mild and transparent assumptions,
structure learning is consistent in high dimensions for
$\threscondalgo$ when the number of samples scales as $n = \Omega
(J_{\min}^{-2}\log p )$, for a $p$-node graph, where $J_{\min}$ is the
minimum (absolute) edge-potential of the Ising model. We relate the
conditions for successful graph recovery to certain phase transitions
in the Ising model. We also derive (nonasymptotic) PAC guarantees for
$\threscondalgo$ and provide explicit results for specific graph
families.

We derive a lower bound (necessary condition) on the sample complexity
required for consistent structure learning with positive probability by
any algorithm. We prove that $n=\Omega(c \log p)$ number of samples is
required by any algorithm to ensure consistent learning of
Erd\H{o}s--R\'{e}nyi random graphs, where $c$ is the average degree,
and $p$ is the number of nodes. We also present a nonasymptotic
necessary condition which employs information-theoretic techniques such
as Fano's inequality and typicality. We also provide results for other
graph families such as the girth-constrained graphs and augmented
graphs.


Our results have several ramifications: we characterize the trade-off
between various graph parameters, such as the maximum degree, threshold
for local path length and the strength of edge potentials for efficient
and consistent structure estimation. For instance, we establish a
natural relationship between maximum degree and girth of a graph for
consistent estimation: graphs with large degrees can be consistently
estimated by our algorithm when they also have large girths. Indeed, in
the extreme case of trees which have infinite girth, they can be
consistently estimated with no constraint on the node degrees,
corroborating the initial observation by Chow and
Liu~\cite{Chow&Liu:68IT}. We also derive stronger guarantees for many
random-graph families. For instance, for the Erd\H{o}s--R\'{e}nyi
random graph family and the small-world family (which is the union of a
$d$-dimensional grid and an Erd\H{o}s--R\'{e}nyi random graph), the
minimum sample complexity scales as $n=\Omega(c^2\log p)$, where $c$ is
the average degree of the Erd\H{o}s--R\'{e}nyi random graph. Thus, when
the average degree is bounded $[c=O(1)]$, the sample complexity of our
algorithm scales as $n =\Omega(\log p)$. Recall that the sample
complexity of learning tree models is $\Omega(\log
p)$~\cite{Tandetald10JMLR}. Thus, we establish that the complexity of
learning sparse random graphs using the proposed algorithm is akin to
learning tree models in certain parameter regimes.

Our sufficient conditions for consistent structure estimation impose
transparent constraints on the graph structure and the parameters. The
structural property is related to the presence of sparse local vertex
separators between nonadjacent node pairs in the graph. The
conditions on the parameters require that the edge potentials of the
Ising model be below a certain threshold, which we explicitly
characterize. In fact, we establish that below this threshold, the
effect of long-range paths in the model decays and that graph
estimation is feasible via local conditioning, as prescribed by our
algorithm. Similar notions have been previously established in other
contexts, for example, to establish polynomial mixing time for Gibbs
sampling of the Ising model~\cite{Levin:book}. We compare these
different criteria and show that we can guarantee consistent learning
in high dimensions under weaker conditions than those required for
polynomial mixing of Gibbs sampling. Ours is the first work (to the
best of the authors' knowledge) to establish such explicit connections
between structure estimation and the statistical physics properties
(i.e., phase transitions) of Ising models. Establishing these results
requires the development and use of tools (e.g., self-avoiding walk
trees), not previously employed for learning problems.





\subsection{Related work}\label{sec:related}

The problem of structure estimation of a general graphical
model~\cite{Kar01,Bogdanov&etal:Rand} is NP-hard. However, for
tree-structured graphical models, the maximum-likelihood (ML)
estimation can be implemented efficiently via the Chow--Liu
algorithm~\cite{Chow&Liu:68IT} since ML estimation reduces to a
maximum-weight spanning tree problem where the edge weights are the
empirical mutual information quantities, computed from samples. It can
be established that the sample complexity for the Chow--Liu algorithm
scales as $n =\Omega(\log p)$, where $p$ is the number of
variables~\cite{Tandetald10JMLR}. Error-exponent analysis of the
Chow--Liu algorithm was performed in
\cite{Tan&etal:09ITsub,Tandetald10SP}, and extensions to general
acyclic models \cite{Tandetald10JMLR,Liudetad10forest} and trees with
latent (or hidden) variables~\cite{Chodetald0JMLR} have also been
studied recently.

Given the feasibility of structure learning of tree models, a natural
extension is to consider learning the structures of {\em junction
trees}.\footnote{Junction trees are formed by triangulating a given
graph, and its nodes correspond to the maximal cliques of the
triangulated graph~\cite{Wainwright&Jordan:08NOW}. The {\em treewidth}
of a graph is one less than the minimum possible size of the maximum
clique in the triangulated graph over all possible triangulations.}
Efficient algorithms have been previously proposed for learning
junction trees with bounded treewidth
(e.g.,~\cite{Chechetkadetald7NIPS}). However, the complexity of these
algorithms is exponential in the tree width, and hence are not
practical when the graphs have unbounded treewidth.\footnote{For
instance, it is known that for a Erd\H{o}s--R\'{e}nyi random graph
$G_p\sim\Gmsc(p, c/p)$ when $(c>1)$, the tree-width is greater than
$p^\varepsilon$, for some $\varepsilon>0$~\cite{Kloks:94LNCS}.}

There are mainly two classes of algorithms for graphical model
selection: local-search based
approaches~\cite{BreslerdetaldRand,SanghavidetadAllerton10} and those
based on convex optimization~\cite
{Mei06,Ravikumardetad08Arxiv,Ravikumardetald08Stat,Chandrasekaran:10latent}.
The latter approach typically incorporates an $\ell_1$ penalty term to
encourage sparsity in the graph structure.
In~\cite{Ravikumardetald08Stat}, structure estimation of Ising models
is considered where neighborhood selection for each node is performed,
based on $\ell_1$-penalized logistic regression. It was shown that this
algorithm has a sample complexity of $n =\Omega(\Delta^3 \log p)$ under
a set of so-called ``incoherence'' conditions. However, the incoherence
conditions are not easy to interpret and NP-hard to verify in general
models~\cite{Bento&Montanari:09NIPS}. For more detailed comparison, see
Section~\ref{sec:previous}.

In contrast to convex-relaxation approaches, the local-search based
approach relies on a series of simple local tests for neighborhood
selection at individual nodes. For instance, the work
in~\cite{BreslerdetaldRand} performs neighborhood selection at each
node based on a series of conditional-independence tests. Abbeel et
al.~\cite{Abbeel2006learning} propose an algorithm, similar in spirit
to learning factor graphs with bounded degree. The authors in
\cite{spirtes1995learning} and \cite{cheng2002learning} consider
conditional-independence tests for learning Bayesian networks. In
\cite{SanghavidetadAllerton10}, the authors suggest an alternative,
greedy algorithm, based on minimizing conditional entropy, for graphs
with large girth and bounded degree. However, these
works~\cite{BreslerdetaldRand,Abbeel2006learning,spirtes1995learning,cheng2002learning,SanghavidetadAllerton10} require the maximum degree
in the graph to be bounded ($\Delta=O(1)$) which may be restrictive in
practical scenarios. We consider graphical model selection on graphs
where the maximum degree is allowed to grow with the number of nodes
(albeit at a controlled rate). Moreover, we establish a natural
trade-off between the maximum degree and other parameters of the graph
(e.g., girth) required for consistent structure estimation.

Necessary conditions on structure learning provide lower bounds on
the sample complexity for structure learning and have been studied in
\cite
{Santhanam&Wainwright:08ISIT,Wang&etal:10ISIT,Mitliagkas&Vishwanath:Allerton10}.
However, a standard assumption that these works make is that the
underlying set of graphs is uniformly distributed with bounded degree.
For this scenario, it is shown that $n=\Omega(\Delta^k \log p)$ samples
are required for consistent structure estimation, for a graph with $p$
nodes and maximum degree $\Delta$, for some $k \in\Nbb$, say $k =3$
or $4$. In contrast, our converse result is stated in terms of the
{\em average degree}, instead of the maximum degree.

\section{System model}
In this section, we define the relevant notation to be used in the rest
of the paper.
\subsection{Notation}
We introduce some basic notions. Let $\norm{\cdot}_1$ denote the
$\ell_1$ norm. For any two discrete distributions $P, Q$ on the same
alphabet $\Xc$, the total variation distance is given by
\begin{equation}\label{eqn:totalvar} \nu(P,Q)
:=\frac{1}{2}\norm{P-Q}_1=\frac{1}{2}\sum_{x\in\calX} |P(x) -
Q(x) |,
\end{equation}
and the Kullback--Leibler distance (or relative entropy) is given
by
\[
D(P\|Q) := \sum_{x\in\calX} P(x)\log\frac{P(x)}{Q(x)}.
\]
Given a pair of discrete random variables $(X,Y)$ taking values on the
set $\calX\times\calY$ and distributed as $P=P_{X,Y}$, the {\em mutual
information} is defined as
\begin{equation}
I(X;Y):=D(P(x,y) \|P(x)P(y))= \sum_{x\in\calX, y\in\calY}
P(x,y)\log
\frac{P(x,y)}{P(x) P(y)}.
\end{equation}
Along similar lines, the {\em conditional mutual information} of $X$
and $Y$ given another random variable $Z$, taking values on a countable
set $\calZ$, is defined as
\begin{equation}
I(X;Y|Z) :=\sum_{x\in\calX, y\in\calY, z\in\calZ} P(x,y,z)\log
\frac{P(x,y|z)}{P(x|z) P(y|z)}.
\end{equation}
It is also well known that $I(X;Y|Z)=0$ if and only if $X$ and $Y$ are
independent given $Z$, that is, $P(x,y|z)=P(x|z)P(y|z)$.

Given $n$ samples drawn i.i.d. from $P(x,y)$, denoted by
$(x^n,y^n)=\{(x_i, y_i)\}_{i=1}^n$, the (joint) {\em empirical
distribution} or the (joint) {\em type} is defined as
\begin{equation}
\hP^n(x,y; x^n,y^n) := \frac{1}{n}\sum_{i=1}^n \Ind\{ (x,y)=(x_i,
y_i)\}.
\end{equation}
%
We loosely use the term {\em empirical distance} to refer to distances
between empirical distributions. For instance, the empirical variation
distance is given by
\begin{equation}\label{eqn:variation}
\nu(\hP^n,\hQ^n)
:=\frac{1}{2}\sum_{x\in\calX} | \hP^n(x)-\hQ^n(x) |.
\end{equation}
Our
algorithm for graph estimation will be based on empirical variation
distance between conditional distributions. We employ such empirical
estimates for testing conditional independencies between specific
distributions.




\subsection{Ising models}

A {\em graphical model} is a family of multivariate distributions which
are Markov in accordance to a particular undirected
graph~\cite{Lauritzen:book}. Each node in the graph $i\in V$ is
associated to a random variable $X_i$, taking value in a set $\Xc$. The
set of edges\footnote{We use the notation $E$ and $G$ interchangeably
to denote the set of edges.} $E\subset\bigl( {{V}\atop{2}}\bigr)$
captures the set of conditional-independence relationships among the
random variables. We say that a vector of random variables
$\bX:=(X_1,\ldots, X_p)$ with a joint probability mass function
(p.m.f.) $P$ is Markov on the graph $G$ if the {\em local Markov
property}
\begin{equation}
P\bigl(x_i|x_{\calN(i)}\bigr) = P(x_i|x_{V\setminus i})
\end{equation}
holds for all nodes $i \in V$. More generally, we say that $P$
satisfies the {\em global Markov property}, if for all disjoint sets
$A,B\subset V$ such that $A\cap\nbd(B)=\nbd(A)\cap B=\varnothing$, we
have
\begin{equation}
P\bigl(\mathbf{x}_A, \mathbf{x}_B|\mathbf{x}_{\Sc(A,
B;G)}\bigr) = P\bigl(\mathbf{x}_A|\mathbf{x}_{\Sc(A, B;G)}\bigr)
P\bigl(\mathbf{x}_B|\mathbf{x}_{\Sc(A, B;G)}\bigr).
\end{equation}
where the set ${\Sc(A,
B;G)}$ is a {\em node separator}\footnote{A set ${\Sc(A, B;G)}\subset
V$ is a separator of sets $A$ and $B$ if the removal of nodes in
${\Sc(A, B;G)}$ separates $A$ and $B$ into distinct components.} between
$A$ and $B$, and $\nbd(A)$ denotes the neighborhood of $A$ in $G$. The
local and global Markov properties are equivalent under the {\em
positivity} condition, given by $P(\mathbf{x})>0$, for all
\mbox{$\mathbf{x}\in\Xc^p$}~\cite{Lauritzen:book}.

The Hammersley--Clifford theorem~\cite{Bremaud:book} states that under
the positivity condition, a distribution $P$ satisfies the Markov
property according to a~graph $G$ if and only if it factorizes
according to the cliques of $G$, that is,
\begin{equation}
P(\mathbf{x}) =
\frac{1}{Z}\exp\biggl(\sum_{c\in\Cc}\Psi_c(\mathbf{x}_c) \biggr),
\end{equation}
where $\Cc$
is the set of cliques of $G$, and $\mathbf{x}_c$ is the set of random
variables on clique $c$. The quantity $Z$ is known as the {\em
partition function} and serves to normalize the probability
distribution. The functions $\Psi_c$ are known as {\em potential}
functions. An important class of graphical models is the class of
pairwise models, which factorize according to the edges of the
graph,
\begin{equation}
P(\mathbf{x}) = \frac{1}{Z}\exp\biggl(\sum_{e\in
E}\Psi_e(\mathbf{x}_e) \biggr).
\end{equation}

One of the most well-studied pairwise models is the Ising model. Here,
each random variable $X_i$ takes values in the set $ \calX=\{-1,+1\}$
and the probability mass function (p.m.f.) is given by
\begin{equation}\label{eqn:Ising}
P({\mathbf x}) =\frac{1}{Z} \exp\biggl[\frac{1}{2}{\mathbf x}^T
\bfJ_G
{\mathbf x}+ \bfh^T {\mathbf x} \biggr],\qquad\mathbf{x}\in\{-1,1\}^p,
\end{equation}
where $\bfJ_G$ is known as the potential matrix, and $\bfh$ as the
potential vector. By convention, $J(i,i)=0$ for all $i \in V$. The
sparsity pattern of $\bfJ_G$ corresponds to that of the graph $G$,
that is,
$J_{i,j}=0$ for $(i,j)\notin G$. A model is said to be {\em attractive}
or {\em ferromagnetic} if $J_{i,j}\geq0$ and $h_i\geq0$, for all
$i,j\in V$. An Ising model is said to be {\em symmetric} if
$\bfh=\bfzero$.

We assume that there exists $ J_{\min}, J_{\max}\in\Rbb$ such that
the absolute values of the edge potentials are uniformly bounded, that
is,
\begin{equation}
|J_{i,j}|\in[J_{\min}, J_{\max}] \qquad\forall(i,j)\in G.
\end{equation}
We can provide guarantees on structure recovery, subject to conditions
on~$J_{\min}$ and~$J_{\max}$. We assume that the node potentials $h_i$
are uniformly bounded away from $\pm\infty$.

Given an Ising model, nodes $i,j\in V$ and a subset $S\subset
V\setminus\{i,j\}$, we define {\em conditional variation distance} as
\begin{eqnarray}\label{eqn:nudef}
\nu_{i|j;S}&:=&\min_{{\mathbf x}_S \in
\{\pm1 \}^{|S|}}\nu\bigl(P(X_i|X_j=+, \bfX_S={\mathbf x}_S),P(X_i|X_j=-,
\bfX_S={\mathbf x}_S)\bigr)\hspace*{-35pt}\\
\label{eqn:condvariation}&=&\min_{{\mathbf x}_S \in\{\pm1\}
^{|S|}}\frac{1}{2}\sum_{x_i=\pm1} | P(X_i=x_i|X_j=+, \bfX
_S={\mathbf
x}_S)\nonumber
\\[-8pt]\\[-8pt]
&&\hphantom{\min_{{\mathbf x}_S \in\{\pm1\}
^{|S|}}\frac{1}{2}\sum_{x_i=\pm1} |}{} - P(X_i=x_i|X_j=-, \bfX_S={\mathbf x}_S)
|.\nonumber
\end{eqnarray}
The empirical conditional variation distance $\hnu_{i|j;S}$ is defined
by replacing the actual distributions with their empirical versions
\begin{eqnarray} \label{eqn:empcondvariation}
\hnu^n_{i,j;S}\!:=\!\min_{{\mathbf x}_S \in\{\pm1\}
^{|S|}}\!\nu\bigl(\hP^n(X_i|X_j\!=\!+, \bfX_S\!=\!{\mathbf x}_S),
\hP^n(X_i|X_j\!=\!-,\bfX_S\!=\!{\mathbf x}_S)\bigr).\hspace*{-40pt}
\end{eqnarray}
Our algorithm will be based on empirical conditional variation
distances. This is because the conditional variation distances\footnote
{Note that the conditional variation distances are in general
asymmetric, that is, $\nu_{i|j;S}\neq\nu_{j|i;S}$.}\vadjust{\goodbreak} can be used as a
test for conditional independence
\begin{equation}
\{X_i \indep X_j |\bfX_S \}
\equiv\{\nu_{i|j;S} =0\} \qquad\forall i,j \in V, S \subset
V\setminus\{i,j\}.
\end{equation}

\subsection{Tractable graph families}

We consider the class of Ising models Mar\-kov on a~graph $G_p$
belonging to some ensemble $\Gmsc(p)$ of graphs with $p$ nodes. We
consider the high-dimensional regime, where both $p$ and the number of
samples $n$ grow simultaneously; typically, the growth of $p$ is much
faster than that of $n$. We emphasize that in our formulation, the
graph ensemble~$\Gmsc(p)$ can either be deterministic or random---in
the latter, we also specify a~probability measure over the set of
graphs in~$\Gmsc(p)$. In the setting where~$\Gmsc(p)$ is a~random-graph
ensemble, let $P_{\bfX, G}$ denote the joint probability distribution
of the variables $\bfX$ and the graph $G\sim\Gmsc(p)$, and let
$P_{\bfX|G}$ denote the conditional distribution of the variables given
a graph $G$. Let $P_G$ denote the probability distribution of graph $G$
drawn from a random ensemble $\Gmsc(p)$. In this setting, we use the
term {\em almost every} (a.e.) graph $G$ satisfies a certain property
$\Qc$ if
\[
\lim_{p\to\infty} P_G[G\mbox{ satisfies }\Qc]=1.
\]
In
other words, the property $\Qc$ holds asymptotically almost
surely\footnote{Note that the term a.a.s. does not apply to
deterministic graph ensembles $\Gmsc(p)$ where no randomness is
assumed, and in this setting, we assume that the property $\Qc$ holds
for every graph in the ensemble.} (a.a.s.) with respect to the
random-graph ensemble $\Gmsc(p)$. Our conditions and theoretical
guarantees will be based on this notion for random graph ensembles.
Intuitively, this means that graphs that have a vanishing probability
of occurrence as $p \to\infty$ are ignored.

\begin{figure}

\includegraphics{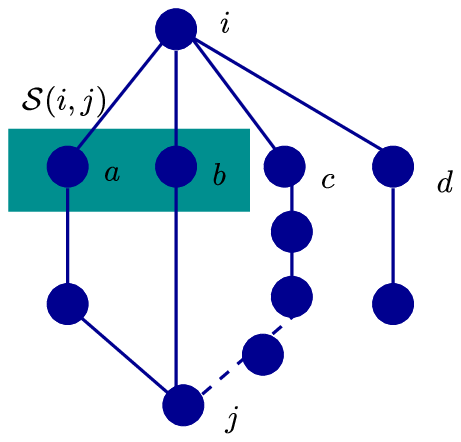}

\caption{Illustration of $l$-local separator set $\calS(i,j;G,l)$
for the graph shown above with $l = 4$. Note that
$\nbd(i)=\{a,b,c,d\}$ is the neighborhood of $i$ and the $l$-local
separator set $\calS(i,j;G,l)=\{a,b\}\subset\nbd(i;G)$. This is
because the path along $c$ connecting $i$ and $j$ has a~length greater
than $l$ and hence node $c\notin\Sc(i,j;G,l)$.}
\label{fig:pseparator}
\end{figure}

We now characterize the ensemble of graphs amenable for consistent
structure estimation under our formulation. To this end, we
characterize the so-called {\em local separators} in graphs. See
Figure~\ref{fig:pseparator} for an illustration. For $\gamma\in\Nbb$, let
$B_\gamma(i;G )$ denote the set of vertices within distance $\gamma$
from $i$ with respect to graph $G$. Let $F_{\gamma,
i}:=G(B_\gamma(i))$ denote the subgraph of $G$ spanned
by~$B_\gamma(i;G)$, but in addition, we retain the nodes not in
$B_\gamma(i)$ (and remove the corresponding edges).

\begin{definition}[($\gamma$-Local separator)]\label{def:sep}
Given
a graph $G$, a
$\gamma$-{\em local separator} $S_\gamma(i,j)$ between $i$ and $j$,
for $(i,j)\notin G$, is a {\em minimal} vertex separator\footnote{A
minimal separator is a separator of smallest cardinality.} with respect
to the subgraph $F_{\gamma,i} $. In addition, the parameter $\gamma$
is referred to as the {\em path threshold} for local separation.
\end{definition}

In other words, the $\gamma$-local separator $S_\gamma(i,j)$ separates
nodes $i$ and $j$ with respect to paths in $G$ of length at most
$\gamma$. We now characterize the ensemble of graphs based on the size
of local separators.\vadjust{\goodbreak}


\begin{definition}[($(\eta,\gamma)$-Local separation property)]\label
{def:localpath}
An ensemble of graphs $\Gmsc(p;\eta,\gamma)$ satisfies
$(\eta,\gamma)$-local separation property if for a.e. $G_p\in
\Gmsc(p;\eta,\gamma)$,
\begin{equation}\label{eqn:localpath}
\max_{(i,j)\notin G_p}|S_\gamma
(i,j)|\leq
\eta.
\end{equation}
\end{definition}

In Section~\ref{sec:method}, we propose an efficient algorithm for
graphical model selection when the underlying graph belongs to a graph
ensemble $\Gmsc(p;\eta, \gamma)$ with sparse local separators [i.e.,
small $\eta$, for $\eta$ defined in \eqref{eqn:localpath}]. We will see
that the computational complexity of our proposed algorithm scales as
$O(p^{\eta+2})$. In Section~\ref{sec:graphexamples}, we provide
examples of many graph families satisfying \eqref{eqn:localpath}, which
include the random regular graphs, Erd\H{o}s--R\'{e}nyi random graphs
and small-world graphs.

\begin{remark*}
The criterion of local separation for tractable learning
is novel to the best of our knowledge. The complexity of a graphical
model is usually expressed in terms of its {\em
tree-width}~\cite{Wainwright&Jordan:08NOW}. We note that the criterion
of sparse local separation is weaker than the tree-width; that is,
$\eta\leq t$, where $t$ is the tree-width of the graph. In fact, our
criterion is also weaker than the criterion of bounded local
tree-width, introduced in~\cite{Epp-Algo-00}.
\end{remark*}

\section{Method and guarantees}\label{sec:method}

\subsection{Assumptions}\label{sec:assume}

\begin{longlist}
\item[(A1)] \textit{Sample complexity}: We consider the asymptotic setting
where both the number of variables (nodes) $p$ and the number of i.i.d.
samples $n$ go to infinity. The required sample complexity is
\begin{equation}
\label{eqn:sample_complexity2}n
=\Omega(J_{\min}^{-2} \log p).
\end{equation}
We require that the number of
nodes $p \to\infty$ to exploit the local-separation properties of the
class of graphs under consideration.

\item[(A2)] \textit{Bounded edge potentials}: The Ising model Markov on
a.e. $G_p\sim\Gmsc(p)$ has the maximum absolute potential below a
threshold $J^*$. More precisely,
\begin{equation}
\label{eqn:uniqueness}\alpha:=\frac{\tanh J_{\max}}{\tanh
J^*}<1,
\end{equation}
where the threshold $J^*$ depends on the specific graph ensemble
$\Gmsc(p)$. See Section~8.1 in the supplementary
material~\cite{AnandkumarTanWillsky:Ising11Supp} for an explicit
characterization of $J^*$ for specific ensembles.

\item[(A3)]\textit{Local-separation property}: We consider the ensemble of
graphs $\Gmsc(p)$ such that almost every graph $G$ drawn from
$\Gmsc(p)$ satisfies the local-separation property $(\eta,\gamma)$,
according to Definition~\ref{def:localpath}, for some $\eta=O(1)$ and
$\gamma\in\Nbb$ such that\footnote{The condition in \eqref{eqn:gamma}
involving $\tilomega(1)$ is required for random graph ensembles such as
Erd\H{o}s--R\'{e}nyi random graphs. It can be weakened as $J_{\min}
\alpha^{-\gamma} = \omega(1)$ for degree-bounded ensembles
$\Gmsc_{\Deg}(\Delta)$.}
\begin{equation}\label{eqn:gamma}
J_{\min} \alpha^{-\gamma} =
\tilomega(1),
\end{equation}
where we say that a function
$f(p)=\tilomega(g(p))$, if $\frac{f(p)}{g(p) \log p}\to\infty$ as
$p\to\infty$.

\item[(A4)]\textit{Generic edge-potentials}: The edge potentials
$\{J_{i,j}, (i,j)\in G\}$ of the Ising model are assumed to be
generically drawn from $[-J_{\max}, -J_{\min}]\cup[J_{\min},
J_{\max}]$; that is, our results hold except for a set of Lebesgue
measure zero. We also characterize specific classes of models where
this assumption can be removed, and we allow for any choice of edge
potentials. See Section~8.3 in the supplementary
material~\cite{AnandkumarTanWillsky:Ising11Supp} for details.
\end{longlist}

Assumption (A1) provides on the bound on the sample complexity.
Assumption (A2) limits the maximum edge potential $J_{\max}$ of the
model. Assumption (A3) relates the path threshold $\gamma$ with the
minimum edge potential $J_{\min}$ in the model. For instance, if
$J_{\min}=\Theta(1)$ and $\gamma=O(\log\log p)$, we require that
$\alpha:=\frac{\tanh J_{\max}}{\tanh J^*}=1-\Theta(1)<1$.

Condition (A4) guarantees the success of our method for generic edge
potentials. Note that if the neighbors are marginally independent,
then our method fails, and thus, we cannot expect our method to succeed
for all edge potentials. Condition (A4) can be removed if we limit to
attractive models (see Section~8.3.1 in the
supplementary material~\cite{AnandkumarTanWillsky:Ising11Supp}), or if
we allow for nonattractive models, but limit to graphs with bounded
local paths (see Section~8.3.3 in the
supplementary material~\cite{AnandkumarTanWillsky:Ising11Supp}). For
general models, we guarantee success of our methods for generic
potentials; that is, we establish that the set of edge potentials where
our method fails has Lebesgue measure zero. Similar assumptions have
been previously employed; for example, in~\cite{kalisch2007estimating}
where learning directed models is considered, it is assumed that the
graphical model is faithful with respect to the underlying graph.

\subsection{Conditional variation distance thresholding}

We now propose an algorithm, termed as conditional variation distance
thresholding ($\threscondalgo$) which is proven to be consistent for
graph reconstruction under the above assumptions. The procedure for
$\threscondalgo$ is provided in Algorithm~\ref{algo:corr_thres_cond}.
Denote $\threscondalgo({\mathbf x}^n;\xi_{n,p})$ as the output edge
set from $\threscondalgo$ given $n$ i.i.d. samples~${\mathbf x}^n$ and
threshold $\xi_{n,p}$. The conditional variation distance test in the
$\threscondalgo$ algorithm computes the empirical conditional
variation distance in \eqref{eqn:empcondvariation} for each node pair
$(i,j)\in V^2$ and finds the conditioning set which achieves the
minimum over all sets of cardinality $\eta$. If the minimum exceeds
the threshold~$\xi_{n,p}$, the node pair is declared an edge.

The threshold $\xi_{n,p}$ needs to separate the edges and the
nonedges in the Ising model. It is chosen as a function of both number
of nodes $p$ and number of samples $n$ and needs to satisfy the
following conditions:
\begin{equation}\label{eqn:xi}\xi_{n,p}= O(J_{\min}),\qquad\xi_{n,p}=
\tilomega(\alpha^{\gamma}),\qquad\xi_{n,p}= \Omega\Biggl(\sqrt{\frac{\log
p}{n} }
\Biggr).
\end{equation}
For example, when $J_{\min}=\Omega(1)$,
$\alpha<1$, $\gamma= \Omega(\log p)$, $n =\Omega( g_p \log p)$, for
some sequence $g_p = \omega(1)$, we can choose $\xi_{n,p}=
\frac{1}{\min(g_p, \log p)}$.

Note that there is dependence on both $n$ and $p$, since we need to
regularize for sample size, as well as for the size of the graph. In
other words, with finite number of samples $n$, the empirical
conditional variation distances are noisy, and the threshold
$\xi_{n,p}$ takes this into account via its inverse dependence on $n$.
Similarly, as the graph size $p$ increases, we establish that the
true conditional variation distance decays at a certain rate under
assumption (A2). Hence the threshold $\xi_{n,p}$ also depends on the
graph size $p$. Moreover, note that for all the conditions in
\eqref{eqn:xi} to be satisfied, the number of samples $n$ should scale
at least at a certain rate with respect to $p$, as given by
\eqref{eqn:sample_complexity2}.

\begin{algorithm}[t]
\begin{algorithmic}
\STATE Initialize $\hG^n_p= (V, \varnothing)$.\STATE For each $i,j
\in
V$, if
\begin{equation}\label{eqn:sep}\mathop{\min_{S\subset V\setminus\{
i,j\}}}_{|S|\leq\eta}\hnu_{i|j;S}> \xi_{n,p},
\end{equation}
then add $(i,j)$ to
$\widehat{G}^n_p$.\\ \STATE Output: $\widehat{G}^n_p$.
\end{algorithmic}
\caption{Algorithm $\threscondalgo({\mathbf x}^n;\xi_{n,p},\eta)$ for
structure learning from ${\mathbf x}^n$ samples based on empirical
conditional variation distances. See~(\protect\ref{eqn:empcondvariation}).}
\label{algo:corr_thres_cond}
\end{algorithm}

\subsubsection{Structural consistency of $\threscondalgo$}

Assuming (A1)--(A4), we have the following result on asymptotic
graph structure recovery.

\begin{theorem}[(Structural consistency of $\threscondalgo$)]\label{thm:corr_thres_cond}
The algorithm
$\threscondalgo$ is consistent for structure recovery of Ising models
Markov on a.e. graph $G_p\sim\Gmsc(p;\eta, \gamma)$:
\begin{equation}
\mathop{\lim_{n,p\to\infty}}_{n=\Omega(J_{\min}^{-2}\log p) }P
[\threscondalgo(\{\mathbf{x}^n\};\xi_{n,p},\eta ) \neq G_p ]=0.
\end{equation}
\end{theorem}

The proof of this theorem is provided in Section~8 in
the supplementary material~\cite{AnandkumarTanWillsky:Ising11Supp}.

\begin{remarks*}
\begin{longlist}
\item[(1)] \textit{Consistency guarantee}:
The $\threscondalgo$ algorithm consistently recovers the structure of
the graphical models, with probability tending to one, where the
probability measure is with respect to both the graph and the
samples. We extend our results and provide finite sample guarantees for
specific graph families in Section~\ref{sec:PAC}. Moreover, if we
require a {\em parameter-free} threshold, that is, we do not know the
exact value of $J_{\min}$ but only its scaling with~$p$, then we need
to choose $\xi_{n,p}= o(J_{\min})$ rather than $\xi_{n,p}=
O(J_{\min})$. In this case, the sample complexity scales as $n =
\omega(J_{\min}^{-2} \log p)$.\vspace*{1pt}

\item[(2)] \textit{Other tests for conditional independence}: We consider
a test based on variation distances. Alternatively other distance
measures can be employed. For instance, it can be proven that the
Hellinger distance and the Kullback--Leibler distance have similar
sample complexity results, while a test based on mutual information has
a worse sample complexity of $\Omega(J_{\min}^{-4}\log p)$ under the
assumptions (A1)--(A4). We term the test based on mutual information as
$\threscmialgo$ and compare its experimental performance with
$\threscondalgo$ in Section~\ref{sec:experiments}.\looseness=1

\item[(3)] \textit{Extension to other models}: The $\threscondalgo$
algorithm can be extended to general discrete models
by considering pairwise variation distance between different
configurations. For instance, we can set
\begin{equation}\nu_{i|j;S}:=\mathop{\sum
_{\lambda_1\neq\lambda_2}}_{\lambda_1,\lambda_2\in\Xc}
\min_{{\mathbf x}_S \in\Xc^{|S|}}\nu\bigl(P(X_i|X_j=\lambda_1, \bfX
_S={\mathbf x}_S),P(X_i|X_j=\lambda_2, \bfX_S={\mathbf x}_S)\bigr).\vspace*{6pt}
\end{equation}
In~\cite{AnandkumarTanWillsky:Gaussian11}, we derive analogous
conditions for Gaussian graphical models. Our approach is also
applicable to models with higher order potentials since it does not
depend on the pairwise nature of Ising models. The conditions for
recovery are based on the notion of {\em conditional uniqueness} and
can be imposed on any model. Indeed the regime of parameters where
conditional uniqueness holds depends on the model and is harder to
characterize for more complex models.
Notice that our algorithm requires only low-order statistics [up to $O(
\eta+2)$] for any class of graphical models which is relevant when we
are dealing with models with higher order potentials.
\end{longlist}
\end{remarks*}

\begin{pf*}{Proof outline} We first analyze the scenario when exact
statistics are available. (i) We establish that for any two
nonneighbors $(i,j)\notin G$, the conditional variation distance in
\eqref{eqn:sep} (based on exact statistics) does not exceed the
threshold $\xi_{n,p}$. (ii) Similarly, we also establish that the
conditional variation distance in \eqref{eqn:sep} exceeds the threshold
$\xi_{n,p}$ for all neighbors $(i,j)\in G$. (iii) We then extend these
results to empirical versions using concentration bounds.
\end{pf*}

\subsubsection{PAC Guarantees for $\threscondalgo$}\label{sec:PAC}

We now provide stronger results for $\threscondalgo$ method in terms of
the probably approximately correct (PAC) model of
learning~\cite{Kearns&Vazirani:book}. This provides additional insight
into the task of graph estimation. Given an Ising model $P$ on graph
$G_p$, recall the definition of conditional variation distance
\[
\nu_{i|j;S}:=\min_{{\mathbf x}_S \in\{-1,+1\}^{|S|}}\nu
\bigl(P(X_i|X_j=+, \bfX_S={\mathbf x}_S),P(X_i|X_j=-, \bfX_S={\mathbf x}_S)\bigr).
\]
Given a graph $G_p$ and $\lambda,\eta>0$, define
\begin{eqnarray}\label{eqn:numindelta}
G'_{p}(V;\lambda)&:=&\Bigl\{ (i,j)\in G_p \dvtx  \mathop{\min_{|S|\leq
\eta}}_{S\subset V\setminus\{i,j\}} \nu_{i|j;S}>\lambda\Bigr\}, \\
\label{eqn:Imax}\nu_{\max}(p;\eta) & :=& \max_{ (i,j)\notin G_p}
\mathop{\min_{|S|\leq\eta}}_{S\subset V\setminus\{i,j\}} \nu_{i|j;S}
.
\end{eqnarray}
For any $\delta> 0$, choose the threshold $\xi_{n,p}$ as
\begin{equation}\xi
_{n,p}(\delta)= \nu_{\max}(p;\eta) +\delta.\label
{eqn:xichoice}
\end{equation}
Define
\begin{equation}P_{\min}:= \mathop{\min_{S\subset
V, |S|\leq\eta+1}}_{ {\mathbf x}= \{\pm1\}^{|S|}} P(\bfX_S =
{\mathbf x}_S).
\end{equation}


\begin{theorem}[(PAC guarantees for $\threscondalgo$)]\label{thm:PAC}
Given an Ising model Markov on graph $G$ and threshold $\xi
_{n,p}(\delta)$ according to \eqref{eqn:xichoice}, $\threscondalgo(\{
{\mathbf x}^n\};\xi_{n,p}(\delta), \eta)$ recovers $G'_{p}(V;\nu
_{\max}(p;\eta)+2\delta)$ for any $\delta>0$, defined in \eqref
{eqn:numindelta}, with probability at least $1-\varepsilon$, when the
number of samples is
\begin{equation}n> \frac{2(\delta+2)^2}{\delta^2 P_{\min}^2}
\biggl[\log\biggl(\frac{1}{\varepsilon} \biggr) + (\eta+2) \log p + (\eta+4) \log2 \biggr],
\end{equation}
and the computational complexity scales as $O(p^{\eta+2})$.
\end{theorem}

\begin{pf}The proof is provided in Section~9
in the
supplementary material~\cite{AnandkumarTanWillsky:Ising11Supp}.
\end{pf}

Thus, the above result characterizes the relationship between the
separation between edges and nonedges\vadjust{\goodbreak} (in terms of conditional
variation distances) and the number of samples required to distinguish
them. A critical parameter in the above result is $\nu_{\max}(p;\eta)$,
the maximum conditional variation distance between nonneighbors. We
now provide nonasymptotic bounds on $\nu_{\max}(p;\eta)$ for specific
graph families satisfying the $(\eta,\gamma)$-local separation
condition. A detailed description of the graph families considered
below is provided in Section~\ref{sec:graphexamples}. On lines of
assumption (A2) in Section~\ref{sec:assume}, define
\begin{equation}
\label{eqn:alphacorr}
\alpha:= \frac{\tanh J_{\max}}{\tanh
J^*}.
\end{equation}
As
we noted earlier, the threshold $J^*$ depends on the graph family. We
characterize both $J^*$ and $\nu_{\max}(p;\eta)$ for various graph
families below.

\begin{lemma}[{[Nonasymptotic bounds on $\nu_{\max}(p;\eta)$ for
graph families]}]\label{corr:samplecomplexity}
The following statements
hold for $\alpha$ in \eqref{eqn:alphacorr}:
\begin{longlist}
\item[(1)] For the degree-bounded ensemble $\Gmsc_{\Deg}(p; \Delta)$,
\begin{equation}\label{eqn:numaxdegree} J^*_{\Deg}=\infty,\qquad \nu
_{\max}(p;\Delta)
=0.
\end{equation}

\item[(2)] For the girth-bounded ensemble $\Gmsc_{\girth}(p;g ,
\Delta)$,
\begin{equation}\label{eqn:numaxgirth}J^*_{\girth}=\atanh\biggl(\frac{1}{\Delta} \biggr),\qquad\nu_{\max}(p;1)
\leq\alpha^{g/2},
\end{equation}
where $\Delta$ is
the maximum degree and $g$ is the girth.

\item[(3)] For the ensemble of $\Delta$-random regular graphs
$\Gmsc_{\reg}(p; \Delta)$,
\begin{equation}J^*_{\reg}=\atanh\biggl(\frac{1}{\Delta}
\biggr).
\end{equation}
Choose any $l\in\Nbb$ such that $l<0.25(0.25 p\Delta+0.5
-\Delta^2)$. Then, with probability at least $1- \Delta^{16l-2}
(p\Delta- 4\Delta^2 - 16l)^{-(8l-1)}$,
\begin{equation}\label{eqn:numaxregular}\nu_{\max}(p;2) \leq
\alpha^{l},
\end{equation}
where $\Delta$ is the degree.

\item[(4)] For the Erd\H{o}s--R\'{e}nyi ensemble $\Gmsc_{\ER}(p, c/p)$,
\begin{equation}J^*_{\ER}=\atanh\biggl(\frac{1}{c} \biggr).
\end{equation}
Choose any $l\in\Nbb$ such
that $l < \frac{\log p}{4\log c}$. When $c>1$, then with probability at
least $1- l e^{\sqrt{125}} p^{-2.5}-l!c^{4l+1}p^{-1}$,
\begin{equation}\label{eqn:numaxER}
\nu_{\max}(p;2) \leq2l^3\alpha^{l}\log p,
\end{equation}
where $c$ is the average degree.

\item[(5)] For the small-world graph ensemble $\Gmsc_{\watts}(p, d,
c/p)$, similar results apply.
\begin{equation}J^*_{\watts}=\atanh\biggl(\frac{1}{c}
\biggr),
\end{equation}
Choose any $l\in\Nbb$ such that $l < \frac{\log p}{4\log c}$.
When $c>1$, with probability at least $1- l e^{\sqrt{125}}
p^{-2.5}-l!c^{4l-1}p^{-1}$,
\begin{equation}\label{eqn:numaxsmallworld}\nu_{\max}(p;d+2)
\leq4l^3\alpha^{l}\log p,
\end{equation}
where $c$ is
the average degree of the Erd\H{o}s--R\'{e}nyi subgraph.
\end{longlist}
\end{lemma}

\begin{pf}
See Corollaries~1 and~2 in
Section~8.1 in the supplementary
material~\cite{AnandkumarTanWillsky:Ising11Supp}.
\end{pf}

Thus, we note that the conditional variation distance is small for
nonneighbors when the maximum edge potential $J_{\max}$ is suitably
bounded. Combining the results above on $\nu_{\max}(p;\eta)$ and the
PAC guarantees in Theorem~\ref{thm:PAC}, we note that a majority of
edges in the Ising model can be learned efficiently under a logarithmic
sample complexity.

\subsection{Examples of tractable graph families}\label{sec:graphexamples}

We now show that the local-separation property in
Definition~\ref{def:localpath} and the assumptions in
Section~\ref{sec:assume} hold for a rich class of graphs.

\begin{example}[(Bounded-degree)]
Any (deterministic or random) ensemble of degree-bounded graphs
$\Gmsc_{\Deg}(p, \Delta)$ satisfies $(\eta,\gamma)$-local separation
property with $\eta=\Delta$ and arbitrary $\gamma\in\Nbb$. This is
because for any node $i \in V$, its neighborhood $\nbd(i)$ exactly
separates it from nonneighbors. Since there is exact separation, we
can establish that the threshold in \eqref{eqn:uniqueness} is infinite
($J^*_{\Deg}=\infty$); that is, there is no constraint on the maximum
edge potential~$J_{\max}$. However, the computational complexity of our
proposed algorithm scales as $O(p^{\Delta+2})$; see
also~\cite{BreslerdetaldRand}. Thus, when $\Delta$ is large, our
proposed algorithm, as well as the algorithm
in~\cite{BreslerdetaldRand}, are computationally intensive. Our goal in
this paper is to relax the bounded-degree assumption and to consider
sequences of ensembles of graph $\Gmsc(p)$ whose maximum degrees may
grow with the number of nodes $p$. To this end, we discuss other
structural constraints which can lead to graphs with sparse local
separators.
\end{example}

\begin{example}[(Bounded local paths)]\label{sec:localpaths}
Another sufficient condition\footnote{For any graph satisfying
$(\eta,\gamma)$-local separation property, the number of
vertex-disjoint paths of length at most $\gamma$ between any two
nonneighbors is bounded above by $\eta$, by appealing to Menger's
theorem for bounded path lengths~\cite{lovasz1978mengerian}. However,
the property of local paths that we describe above is a stronger notion
than having sparse local separators, and we consider all distinct
paths of length at most $\gamma$ and not just vertex disjoint paths in
the formulation.} for the $(\eta,\gamma)$-local separation property in
Definition~\ref{def:localpath} to hold is that there are at most
$\eta$ paths of length at most $\gamma$ in $G$ between any two nodes
[henceforth, termed as the $(\eta,\gamma)$-{\em local paths property}].
In other words, there are at most $\eta-1$ number of
overlapping\footnote{Two cycles are said to overlap if they have common
vertices.} cycles of length smaller than $2\gamma$. We denote this
ensemble of graphs as $\Gmsc_{\LP}(p;\eta, \gamma)$.

In particular, a special case of the local-paths property described
above is the so-called girth property. The {\em girth} of a graph is
the length of the shortest cycle. Thus, a graph with girth $g$
satisfies $(\eta, \gamma)$-local separation property with $\eta=1$ and
$\gamma=g/2$. Let $\Gmsc_{\girth}(p;g)$ denote the ensemble of graphs
with girth at most $g$. There are many graph constructions which lead
to large girth. For example, the bipartite Ramanujan
graph~\cite{Chung:book2}, page 107 and the random Cayley
graphs~\cite{gamburd2009girth} have large girths. Recently, efficient
algorithms have been proposed to generate large girth graphs
efficiently~\cite{bayati2009generating}.

The girth condition can be weakened to allow for a small number of
short cycles, while not allowing for typical node neighborhoods to
contain short cycles. Such graphs are termed as {\em locally
tree-like}. For instance, the ensemble of Erd\H{o}s--R\'{e}nyi graphs
$\Gmsc_{\ER}(p, c/p)$, where an edge between any node pair appears with
a probability $c/p$, independent of other node pairs, is locally
tree-like. The parameter $c$ may grow with $p$, albeit at a
controlled rate for tractable structure learning, made precise later.
In Section~11 in the supplementary
material~\cite{AnandkumarTanWillsky:Ising11Supp}, we establish that
there are at most two paths of length smaller than $\gamma< \frac
{\log
p}{4\log c}$ between any two nodes in Erd\H{o}s--R\'{e}nyi graphs
a.a.s., or equivalently, there are no overlapping cycles of length
smaller than $2 \gamma$ a.a.s. Similar observations apply for the more
general {\em scale-free} or {\em power-law}
graphs~\cite{Chung:book,dommers2010ising}, and we derive the precise
relationships in Section~11 in the supplementary
material~\cite{AnandkumarTanWillsky:Ising11Supp}. Along similar lines,
the ensemble of $\Delta$-random regular graphs, denoted by
$\Gmsc_{\reg}(p, \Delta)$, which is the uniform ensemble of regular
graphs with degree $\Delta$ has no overlapping cycles of length at
most $\Theta(\log_{\Delta-1} p)$ a.a.s.~\cite{mckay2004short}, Lemma 1.

We now discuss the conditions under which a general local-paths graph
ensemble $\Gmsc_{\LP}(p;\eta, \gamma)$ satisfies
assumption\footnote{In
fact, a weaker version of (A3) as $J_{\min} \alpha^{-\gamma} =
\omega(1)$ suffices for degree-bounded ensembles
$\Gmsc_{\Deg}(\Delta)$.} (A3) in Section~\ref{sec:assume}, required
for our graph estimation algorithm $\threscondalgo$ to succeed. Denote
the maximum degree for the $\Gmsc_{\LP}(p;\eta, \gamma)$ ensemble as
$\Delta$ (possibly growing with $p$). Note that we can now implement
the $\threscondalgo$ algorithm with parameter $\eta$. In
Section~8.1 in the supplementary
material~\cite{AnandkumarTanWillsky:Ising11Supp}, we establish that
the threshold $J^*$ in \eqref{eqn:uniqueness} is given by\vadjust{\goodbreak}
$J^*_{\LP}=\Theta(1/\Delta)$. When the minimum edge potential
$J_{\min}$ achieves the bound, that is, $J_{\min}=\Theta(1/\Delta)$,
the assumption (A3) simplifies as
\begin{equation}\label{eqn:condDelta}\Delta
\alpha^{\gamma} = o(1).
\end{equation}
Note that $\alpha<1$ under (A2). We obtain
a natural trade-off between the maximum degree $\Delta$ and the path
threshold $\gamma$.

When $\Delta=O(1)$, we can allow the path threshold in
\eqref{eqn:condDelta} to scale as $\gamma= O(\log\log p)$. This
implies that graphs with fairly small path threshold~$\gamma$ can be
incorporated under our framework. In particular, this includes the
class of girth-bounded graph with fairly small girth [i.e., the girth
$g$ scaling as $O(\log\log p)$].

We can also incorporate graph families with growing maximum degrees
in~\eqref{eqn:condDelta}. For instance, when $\Delta=O(\poly\log p)$, we
require the path threshold to scale as $\gamma=O(\log p)$. In
particular, the $\Delta$-random-regular ensemble
satisfies~\eqref{eqn:condDelta} when $\Delta=O(\poly\log p)$.

Thus, \eqref{eqn:condDelta} represents a natural trade-off between node
degrees and path threshold for consistent structure estimation; graphs
with large degrees can be learned efficiently if their path thresholds
are large. Indeed, in the extreme case of trees which have infinite
threshold (since they have infinite girth), in accordance with
\eqref{eqn:condDelta}, there is no constraint on node degrees for
successful recovery, and recall that the Chow--Liu
algorithm~\cite{Chow&Liu:68IT} is an efficient method for model
selection on tree distributions.

Moreover, the constraint in \eqref{eqn:condDelta} can be weakened for
random graph ensembles by replacing the maximum degree with the average
degree. Recall that in the Erd\H{o}s--R\'{e}nyi ensemble
$\Gmsc_{\ER}(p, c/p)$, an edge between any two nodes occurs with
probability $c/p$ and that this ensemble satisfies the $(\eta,\gamma)$
property with path threshold $\gamma=O(\frac{\log p}{\log c})$ and
$\eta=2$. In Section~8.1 in the supplementary
material~\cite{AnandkumarTanWillsky:Ising11Supp}, we establish that
the threshold in \eqref{eqn:uniqueness} is given by
$J^*_{\ER}=\Theta(1/c)$. Comparing with the threshold for
$\Delta$-degree bounded graphs $J^*=\Theta(1/\Delta)$ discussed above,
we see that we can obtain better bounds for random-graph ensembles.


When the minimum edge potentials achieves the threshold
($J_{\min}=\Theta(1/c)$), the requirement in assumption (A3) in
Section~\ref{sec:assume} simplifies to
\begin{equation}\label{eqn:ERA3} c
\alpha^{\gamma}=\tilo(1),
\end{equation}
which is true when $c=O(\poly\log p)$. Thus, we can guarantee
consistent structure estimation for the Erd\H{o}s--R\'{e}nyi ensemble
when the average degree scales as $c=O(\poly\log p)$. This regime is
typically known as the ``sparse'' regime and is relevant, since in
practice, our goal is to fit the measurements to a sparse graphical model.
\end{example}

\begin{example}[(Small-world graphs)]
The previous two examples showed that local separation holds under two
different conditions: bounded maximum degree and bounded number of
local paths. The former class of graphs can have short cycles, but the
maximum degree needs to be constant, while the latter class of graphs
can have a large maximum degree but the number of overlapping short
cycles needs to be small. We now provide instances which incorporate
both these features, large degrees and short cycles, and yet satisfy
the local separation property.

The class of hybrid graphs or augmented graphs~(\cite{Chung:book}, Chapter~12) consists of graphs which are the union of two graphs:
a ``local'' graph, having short cycles, and a ``global'' graph, having
small average distances. Since the hybrid graph is the union of these
local and global graphs, it simultaneously has large degrees and short
cycles. The simplest model $\Gmsc_{\watts}(p, d, c/p)$, first studied
by Watts and Strogatz~\cite{watts1998collective}, consists of the union
of a $d$-dimensional grid and an Erd\H{o}s--R\'{e}nyi random graph
with parameter $c$. It is easily seen that a.e. graph
$G\sim\Gmsc_{\watts}(p, d, c/p)$ satisfies $(\eta,\gamma)$-local
separation property in \eqref{eqn:localpath}, with
\[
\eta=d+2,\qquad \gamma\leq\frac{\log p}{4\log c}.
\]
Similar observations apply for
more general hybrid graphs studied in~\cite{Chung:book}, Chapter~12.


In Section~8.1 in the supplementary material, we
establish that the threshold in \eqref{eqn:uniqueness} for the
small-world ensemble $\Gmsc_{\watts}(p, d, c/p)$ is given by $
J^*_{\watts}=\Theta(1/c)$ and is independent of $d$, the degree of the
grid graph. Comparing with the threshold $J^*_{\ER}$ for
Erd\H{o}s--R\'{e}nyi ensemble $\Gmsc_{\ER}(p, c/p)$, we note that the
two thresholds are identical. This further implies that
\eqref{eqn:ERA3} holds for the small-world graph ensemble as well.
\end{example}

\subsection{Explicit bounds on sample complexity of $\threscondalgo$}


Recall that the sample complexity of the $\threscondalgo$ is required
to scale as $n=\Omega(J_{\min}^{-2} \log p)$ for structural
consistency in high dimensions. Thus, the sample complexity is small
when the minimum edge potential $J_{\min}$ is large. On the other
hand, $J_{\min}$ cannot be arbitrarily large due to assumption (A2) in
Section~\ref{sec:assume}, which entails that $J_{\min}<J^*$. The
minimum sample complexity is thus attained when $J_{\min}$ achieves
the threshold~$J^*$.

We now provide explicit results for the minimum sample complexity for
various graph ensembles, based on the threshold $J^*$. Recall that in
Section~\ref{sec:graphexamples}, we discussed that for the graph
ensemble $\Gmsc_{\LP}(p, \eta,\gamma, \Delta)$ satisfying the
$(\eta,\gamma)$-local paths property and having maximum degree
$\Delta$, the threshold is $J^*_{\LP}=1/\Delta$. Thus, the minimum
sample complexity for this graph ensemble is $n = \Omega(\Delta^2
\log
p)$, that is, when $J_{\min}=\Theta(1/\Delta)$.

For the Erd\H{o}s--R\'{e}nyi random graph ensemble $\Gmsc_{\ER}(p,
c/p)$ and the small-world graph ensemble $\Gmsc_{\watts}(p,d,c/p)$,
recall that the thresholds are given by $J^*_{\ER}= J^*_{\watts}=1/c$,
where $c$ is the mean degree of the Erd\H{o}s--R\'{e}nyi graph. Thus,
the minimum sample complexity can be improved to $n =\Omega(c^2\log
p)$, by setting $J_{\min}=\Theta(1/c)$. This implies that when the
Erd\H{o}s--R\'{e}nyi random graphs and small-world graphs have a
bounded average degree $[c=O(1)]$, the minimum sample complexity is
$n=\Omega(\log p)$. Recall that the sample complexity of learning tree
models is $\Omega(\log p)$~\cite{Tandetald10JMLR}. Thus, we observe
that the complexity of learning sparse Erd\H{o}s--R\'{e}nyi random
graphs and small-world graphs using our algorithm $\threscondalgo$ is
akin to learning tree structures in certain parameter regimes.

\subsection{Comparison with previous results}\label{sec:previous}

We now compare the performance of our algorithm $\threscondalgo$ with
$\ell_1$-penalized logistic regression proposed
in~\cite{Ravikumardetald08Stat}. We first compare the computational
complexities. The method in~\cite{Ravikumardetald08Stat} has a
computational complexity of $O(p^4)$ for any input (assuming $p>n$). On
the other hand, the complexity of our method depends on the graph
family under consideration. It can be as low as $O(p^3)$ for
girth-bounded ensembles, $O(p^4)$ for random graph families and as high
as $O(p^\Delta)$ for degree-bounded ensembles (without any additional
characterization of the local separation property). Clearly our method
is not efficient for general degree-bounded ensembles since it is
tailored to exploit the sparse local-separation property in the
underlying graph.

We now compare the sample complexities under the two methods. It was
established that the method in~\cite{Ravikumardetald08Stat} has a
minimum sample complexity of $n = \Omega(\Delta^3\log p)$ for a
degree-bounded ensemble $\Gmsc_{\Deg}(p,\Delta)$ satisfying certain
``incoherence'' conditions. The sample complexity of our
$\threscondalgo$ algorithm is better at $n=\Omega(\Delta^2 \log p)$.
Moreover, we can guarantee improved sample complexity of $n
=\Omega(c^2\log p)$ for Erd\H{o}s--R\'{e}nyi random graphs
$\Gmsc_{\ER}(p,c/p)$ and small-world graphs $\Gmsc_{\watts}(p, d, c/p)$
under the modified $\threscondalgo$ algorithm. Note that these random
graph ensembles have maximum degrees ($\Delta$) much larger than the
average degrees ($c$), and thus, we can provide stronger sample
complexity results. Moreover, our algorithm is local and requires only
low-order statistics for any class of graphical models of arbitrary
order, while the method in~\cite{Ravikumardetald08Stat} requires
full-order statistics since it undertakes neighborhood selection
through regularized logistic regression. This is relevant in practice,
since our algorithm is better equipped to handle missing samples.


The incoherence conditions required for the success of $\ell_1$
penalized logistic regression in~\cite{Ravikumardetald08Stat} are
NP-hard to establish for general models since they involve the
partition function of the model~\cite{Bento&Montanari:09NIPS}. In
contrast, our conditions are transparent and relate to the phase
transitions in the model. It is an open question as to whether the
incoherence conditions are implied by our assumptions or vice-versa for
general models. It appears that our conditions are weaker than the
incoherence conditions for random-graph models. For instance, for the
Erd\H{o}s--R\'{e}nyi model $\Gmsc_{\ER}(p, c/p)$, we require that
$J_{\max} = O(1/c)$, where $c$ is the average degree, while a
sufficient condition for incoherence is $J_{\max}=O(1/\Delta)$, where
$\Delta$ is the maximum degree. Note that\vadjust{\goodbreak} $\Delta=O(\log p \log c)$
a.a.s. for the Erd\H{o}s--R\'{e}nyi model. Similar observations also
hold for the power-law and small-world graph ensembles. This implies
that we can guarantee consistent structure estimation under weaker
conditions (i.e., a wider range of parameters) and better sample
complexity for the Erd\H{o}s--R\'{e}nyi, power-law and small-world
models.


\section{Necessary conditions for graph estimation}\label{sec:lowerbound}

We have so far proposed algorithms and provided performance
guarantees for graph estimation given samples from an Ising models.
We now analyze necessary conditions for graph estimation.

\subsection{Erd\H{o}s--R\'{e}nyi random graphs}

Necessary conditions for graph estimation have been previously
characterized for degree-bounded graph ensembles $\Gmsc_{\Deg}(p,
\Delta)$~\cite{Santhanam&Wainwright:08ISIT}. However, these conditions
are too loose to be useful for the ensemble of Erd\H{o}s--R\'{e}nyi
graphs $\Gmsc_{\ER}(p,c/p)$, where the average degree\footnote{The
techniques in this section is applicable when the average sparsity
parameter $c$ of $\Gmsc_{\ER}(p,c/p)$ ensemble is a function of $p$
and satisfies $c \leq p/2$.} $(c)$ is much smaller than the maximum
degree.

We now provide a lower bound on sample complexity for graph estimation
of Erd\H{o}s--R\'{e}nyi graphs using any deterministic estimator.
Recall that $p$ is the number of nodes in the model, and $n$ is the
number of samples. In the following result, $c$ is allowed to depend on
$p$ and is thus more general than the previous results.

\begin{theorem}[(Necessary conditions for model selection)]\label{thm:lowerbound}
Assume that $c\le0.5p$ and $G_p \sim\Gmsc_{\ER}
(p,c/p)$. Then if $n \le\varepsilon c \log p$ for sufficiently small
$\varepsilon>0$, we have
\begin{equation}
\lim_{p\to\infty} P[\hG^n_p(\bfX^n_p) \ne G_p]= 1
\end{equation}
for any deterministic estimator $\hG_p$.
\end{theorem}

Thus, when $n \le\varepsilon c \log p$ for sufficiently small
$\varepsilon>0$, the probability of error for structure estimation
tends to one, where the probability measure is with respect to both the
Erd\H{o}s--R\'{e}nyi random graph and the samples. The proof of this
theorem can be found in Section~10 in the
supplementary material, and is along the lines of~\cite{BreslerdetaldRand}, Theorem 1.

The result in Theorem~\ref{thm:lowerbound} provides an asymptotic
necessary condition for structure learning and involves an additional
auxiliary parameter $\varepsilon$. In the following result, we remove
the requirement for the auxiliary parameter $\varepsilon$ and provide a
nonasymptotic necessary condition, but at the expense of having a weak
(instead of a strong) converse.

%
\begin{theorem}[(Nonasymptotic necessary conditions for model
selection)]\label{thm:weakconverse}
Assume that $G\sim \mathcal{G}_{\mathrm{ER}}(p,c/p)$,\vadjust{\goodbreak} where $c$ may depend on $p$. Let
$P_e^{(p)} :=\break P(\hat{G}_p \ne G_p)$ be the probability of error. If
$P_e^{(p)}\to0$, the number of samples~$n$ must satisfy
\begin{equation}\label{eqn:fano}
n\ge\frac{1}{p\log_2|\calX|}\pmatrix{p\cr 2} \Hc_b\biggl( \frac{c}{p} \biggr).
\end{equation}
\end{theorem}
By expanding the binary entropy function $\Hc_b(\cdot)$, it is easy to
see that the statement in (\ref{eqn:fano}) can be weakened to the more
easily interpretable (albeit weaker) necessary condition
\begin{equation}
n\ge\frac{c\log_2 p}{2\log_2|\calX|}.
\end{equation}

The above result differs from Theorem~\ref{thm:lowerbound} in two
aspects: the bound in~\eqref{eqn:fano} does not involve any
asymptotic notation and is a weak converse result (instead of a strong
converse). The proof is provided in
Section~10.3 in the supplementary
material~\cite{AnandkumarTanWillsky:Ising11Supp}.

\begin{remarks*}
\begin{longlist}
\item[(1)] Thus, $n=\Omega(c\log p)$ number of samples are
\emph{necessary} for structure recovery. Hence, the larger the
average degree, the higher is the required sample complexity.
Intuitively this is because as $c$ grows, the graph is denser, and
hence we require more samples for learning. In information-theoretic
terms, Theorem~\ref{thm:lowerbound} is a strong
converse~\cite{Cover&Thomas:book}, since we show that the error
probability of structure learning tends to one (instead of being merely
bounded away from zero). On the other hand, the result in
Theorem~\ref{thm:weakconverse} is a weak converse result.
\item[(2)] In \cite{Santhanam&Wainwright:08ISIT}, it is shown that for
graphs uniformly drawn from the class of graphs with maximum degree
$\Delta$, when $n < \varepsilon\Delta^k \log p$ for some $k\in\Nbb$,
there exists a graph for which any estimator fails with probability at
least $0.5$. These results cannot be applied here since the probability
mass function is nonuniform for the class of Erd\H{o}s--R\'{e}nyi
random graphs.
%
\item[(3)] The result is not dependent on the Ising model assumption,
and holds for {\em any} pairwise discrete Markov random field (i.e.,
$\Xc$ is a finite set).
\end{longlist}
\end{remarks*}

We now provide an outline for the proof of
Theorem~\ref{thm:weakconverse}. A na\"{i}ve application of Fano's
inequality for this problem does not yield any meaningful result since
the set of all graphs (which can be realized by $\Gmsc_{\ER}$) is ``too
large.'' We employ another information-theoretic idea known as {\em
typicality}. We identify a set of graphs with $p$ nodes whose average
degree is $\varepsilon$-close to $c$ (which is the expected degree for
$\Gmsc_{\ER}(p,c/p)$. The set of typical graphs has a small cardinality
but high probability when $p$ is large. The novelty of our proof lies
in our use of both typicality as well as Fano's inequality to derive
necessary conditions for structure learning. We can show that (i) the
probability of the typical set tends to one as $p\to\infty$; (ii) the
graphs in the typical set are almost uniformly distributed (the
asymptotic equipartition property); (iii)~the cardinality of the
typical set is small relative to the set of all graphs. A~detailed
discussion of these techniques is given
in~\cite{AnandkumarTanWillsky:Gaussian11}.


\subsection{Other graph families}

We now provide necessary conditions for recovery of graphs belonging to
various graph ensembles considered in this paper. We first recap the
results of~\cite{BreslerdetaldRand}, Theorem 1, which is applicable for
any uniform ensemble of graphs.

\begin{theorem}[(Lower bound on sample complexity)]\label{thm:lowerbounduniform}
$\!\!\!$Assume that a graph~$G_p$ on $p$ nodes is
uniformly drawn from an ensemble $\Gmsc$. Given $n$ i.i.d. samples
from an Ising model Markov on $G$, we have
\begin{equation}
P[\hG^n_p(\bfX^n_p) \ne G_p]\geq1-\frac{2^{np}}{|\Gmsc|}
\end{equation}
for any deterministic estimator $\hG_p$.
\end{theorem}

We provide bounds on the number of graphs in specific graph families
considered earlier in the paper which gives us necessary conditions for
their recovery.

\begin{lemma}[(Bounds on size of graph families)]\label{lemma:numgraphs}
The following bounds hold:
\begin{longlist}
\item[(1)] For girth-bounded ensembles $\Gmsc_{\girth}(p;g, \Delta
_{\min},\Delta_{\max},k)$ with girth $g$, minimum degree $\Delta
_{\min}$, maximum degree $\Delta_{\max}$ and number of edges $k$, we have
\begin{equation}p^k (p-g\Delta_{\max}^g)^k \leq|\Gmsc_{\girth}(p;g,
\Delta_{\min}, \Delta_{\max},k)|\leq p^k (p-\Delta_{\min
}^g)^k.
\end{equation}

\item[(2)] For local-path ensembles $\Gmsc_{\LP}(p;\eta, \gamma,
\Delta_{\min}, \Delta_{\max}, k)$ having $\eta$ paths of length less
than $\gamma>0$ between any two nodes, minimum degree
$\Delta_{\min}>0$, maximum degree $\Delta_{\max}$ and number of edges
$k$,
\begin{eqnarray}   m_1 p^{k_1} (p-\gamma\Delta_{\max}^\gamma)^{k_1}
\pmatrix{\Delta_{\min}^\gamma\cr2}^{\eta-1}&\leq&|\Gmsc_{\LP
}(p;\eta,
\gamma, \Delta_{\min}, \Delta_{\max}, k)|\nn\\ &\leq&
m_2p^{k_2}(p-\Delta_{\min}^\gamma)^{k_2}
\pmatrix{\gamma\Delta_{\max}^\gamma\cr2}^{\eta-1},
\end{eqnarray}
where
$k_1:=k-m_2(\eta-1)$, $k_2:=k -m_1(\eta-1)$, $m_1:=\frac{p}{\gamma
\Delta_{\max}^\gamma}$ and $m_2:=\frac{p}{ \Delta_{\min}^\gamma}$.

\item[(3)] For augmented ensembles $\Gmsc_{\Aug}(p;d,\eta, \gamma,
\Delta_{\min}, \Delta_{\max}, k)$ consisting of a local graph with
(regular) degree $d$ and a global graph\vadjust{\goodbreak} $\Gmsc_{\LP}(p;\eta, \gamma,
\Delta_{\min},\break \Delta_{\max}, k)$, we have
\begin{eqnarray}   &&m_1 p^{k'_1} (p-\gamma\Delta_{\max}^\gamma)^{k'_1}
\pmatrix{\Delta_{\min}^\gamma\cr2}^{\eta-1}\pmatrix{p-1\cr d}\nonumber
\\
&&\qquad\leq
|\Gmsc_{\Aug}(p;d,\eta, \gamma, \Delta_{\min}, \Delta_{\max},
k)|\\
&&\qquad\leq m_2p^{k'_2}(p-\Delta_{\min}^\gamma)^{k'_2}
\pmatrix{\gamma\Delta_{\max}^\gamma\cr2}^{\eta-1}\pmatrix{p-1\cr d},\nonumber
\end{eqnarray}
where $k'_1:=k_1+1-\frac{pd}{2}$ and $k'_2:=k_2+1-\frac{pd}{2}$, for
$k_1, k_2, m_1, m_2$ defined previously.
\end{longlist}
\end{lemma}

The proof of the above result is given in
Section~10.2 in the supplementary material \cite{AnandkumarTanWillsky:Ising11Supp}.

\begin{remarks*}
Using the above results on lower bounds on the number of
graphs in a given family, in conjunction with
Theorem~\ref{thm:lowerbounduniform}, we can obtain necessary conditions
for different graph families. For instance, for girth-constrained
families, when the girth $g$ and maximum degree $\Delta_{\max}$ scale
as $O(\poly\log p)$, we have that
\begin{equation}\label{eqn:nensemble} n = \Omega
\biggl[\frac{k}{p}\log p \biggr]
\end{equation}
number of samples is necessary for structure
estimation, where $k$ is the number of edges. Similarly, for local path
ensembles, when the path threshold~$\gamma$ and maximum degree
$\Delta_{\max}$ scale as $O(\poly\log p)$, the above bound in
\eqref{eqn:nensemble} changes only slightly, and we have
\[
n = \Omega\biggl[ \biggl(\frac{k}{p}-\frac{\eta-1}{\Delta_{\min}^\gamma}
\biggr)\log p \biggr]
\]
as the necessary condition, by substituting for
$k_1$, and noting that the other terms scale slower than $\log p$ under
the above specified regime. Similarly, for augmented graphs, we have
\[
n = \Omega\biggl[ \biggl(\frac{k}{p}-\frac{\eta-1}{\Delta_{\min}^\gamma}
-\frac{d}{2} \biggr)\log p \biggr]
\]
as the necessary condition. Thus,
for a wide class of graphs, we can characterize necessary conditions
for structure estimation.
\end{remarks*}

\section{Experiments}\label{sec:experiments}

In this section experimental results are presented on synthetic data.
We implement the proposed $\threscondalgo$ (based on conditional
variation distances) and $\threscmialgo$ (based on conditional mutual
information) methods under\vadjust{\goodbreak} different thresholds, as well the $\ell_1$
regularized logistic regression~\cite{Ravikumardetald08Stat} under
different regularization parameters.\footnote{For the convex relaxation
method in~\cite{Ravikumardetald08Stat}, the regularization parameter
denotes the weight associated with the $\ell_1 $ term.} The
performance of the methods is compared using the notion of the edit
distance between the estimated and the true graphs. We implement the
proposed $\threscondalgo$ and $\threscmialgo$ methods in MATLAB and the
$\ell_1$ regularized logistic regression is evaluated using L1General
package.\footnote{L1General is available at
\texttt{\href{http://www.di.ens.fr/\textasciitilde mschmidt/Software/L1General.html}{http://www.di.ens.fr/\textasciitilde mschmidt/Software/L1General.}
\href{http://www.di.ens.fr/\textasciitilde mschmidt/Software/L1General.html}{html}}.}
CONTEST\footnote{CONTEST is at
\texttt{\href{http://www.mathstat.strath.ac.uk/research/groups/numerical\_analysis/contest}{http://www.mathstat.strath.ac.uk/research/groups/numerical\_}
\href{http://www.mathstat.strath.ac.uk/research/groups/numerical\_analysis/contest}{analysis/contest}}.}
package is used to generate the synthetic graphs, and UGM\footnote{UGM
is at \url{http://www.di.ens.fr/\textasciitilde mschmidt/Software/UGM.html}.} package
is used for implementing Gibbs sampling from the Ising Model. The
datasets, software code and results are available at \url{http://newport.eecs.uci.edu/anandkumar}.

%

\subsection{Data sets}
In order to evaluate the $\threscondalgo$ performance in terms of
quantity of errors in recovering the graph structure, we generate
samples from Ising model for three typical graphs, namely, a single
cycle graph whose $\eta_{\mathrm{cycle}}=2$,
Erd\H{o}s--R\'{e}nyi random graph $ \mathcal{G}_{\mathrm{ER}}(p,c/p)$ with
average degree $c=1$ and the Watts and Strogatz model
$\mathcal{G}_{\mathrm{WS}}(p,d,c/p)$ with degree of local graph $d=2$ and
average degree of the global graph $c=1$. Graphs of size $p=80$ and
sample size $n\in\{10^2,5\times10^2,10^3,5\times10^3,10^4,10^5\}$ are
considered.

\begin{table}
\tabcolsep=0pt
\caption{Normalized edit distance under $\threscondalgo$ (based on
conditional variation distances), $\threscmialgo$ (based on
conditional mutual information) and $\ell_1$ penalized neighborhood
selection on synthetic data from graphs listed above for attractive and
mixed Ising models, where $n$ denotes the number of samples}\label{table:collectionthreshold}
\begin{tabular*}{\textwidth}{@{\extracolsep{\fill}}lccccccc@{}}
\hline
\textbf{Graph} & $\bolds{n}$ & $\bolds{\threscondalgo}$ & $\bolds{\threscmialgo}$ &$\bolds{\ell_1}$ \textbf{penalty} &
$\bolds{\threscondalgo}$ & $\bolds{\threscmialgo}$ &$\bolds{\ell_1}$ \textbf{penalty}\\ & &
\textbf{(attractive)} & \textbf{(attractive)}& \textbf{(attractive)} & \textbf{(mixed)} & \textbf{(mixed)} &
\textbf{(mixed)}\\
\hline
Cycle & $1\times10^2$& 1.0000 & 1.0000&1.0000& 1.0000 & 1.0000&1.0000\\
ER & $1\times10^2$ & 1.0000 & 1.0000&1.0000& 1.0000 & 1.0000&1.0000\\
WS & $1\times10^2$ & 1.0000 & 1.0000&1.0000& 1.0000 & 1.0000&1.0000\\[5pt]
Cycle & $5\times10^2$ & 1.0000& 0.5000 &1.0000&0.975\phantom{0}&0.475\phantom{0}& 1.0000\\
ER & $5\times10^2$ & 1.0000 & 0.5300&1.0000&0.9189&0.5946& 1.0000\\
WS & $5\times10^2$ & 1.0000 & 0.3313&1.0000&1.0000&0.3313& 1.0000\\[5pt]
Cycle & $1\times10^3$ & 0.7125 & 0.1750&0.4000&0.7250&0.1500&0.3063\\
ER & $1\times10^3$ & 0.7428&0.1020&0.3378&0.6757&0.1351&0.4342\\
WS & $1\times10^3$ & 0.9937&0.1438&0.1625&0.9938&0.1438&0.4255\\[5pt]
Cycle & $5\times10^3$ & 0.0125&0.0000&0.1937&0.0125&0.0000&0.1500\\
ER & $5\times10^3$ & 0.0000&0.0204&0.2031&0.0000&0.1053&0.0000\\
WS & $5\times10^3$ & 0.3827&0.0000&0.0312&0.5688&0.0000&0.2671\\[5pt]
Cycle & $1\times10^4$ & 0.0000& 0.0000& 0.0000& 0.3063& 0.0000&
0.0000\\
ER & $1\times10^4$ & 0.0000& 0.0000& 0.0000& 0.0000& 0.0000& 0.0000\\
WS & $1\times10^4$ & 0.0000& 0.0000& 0.0000 & 0.0000& 0.0000& 0.0000\\
\hline
\end{tabular*}
\end{table}

\begin{figure}
\begin{tabular*}{\textwidth}{@{\extracolsep{\fill}}cc@{}}

\includegraphics{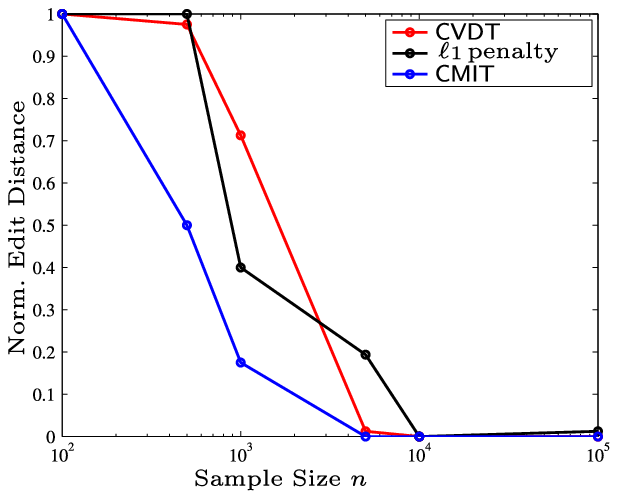}
&\includegraphics{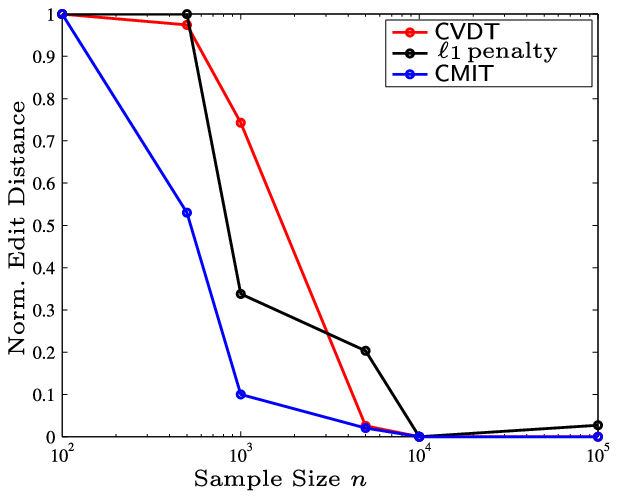}\\
(a) Cycle&(b) Erd\H{o}s-R\'{e}nyi
\\[4pt]
\multicolumn{2}{c}{
\includegraphics{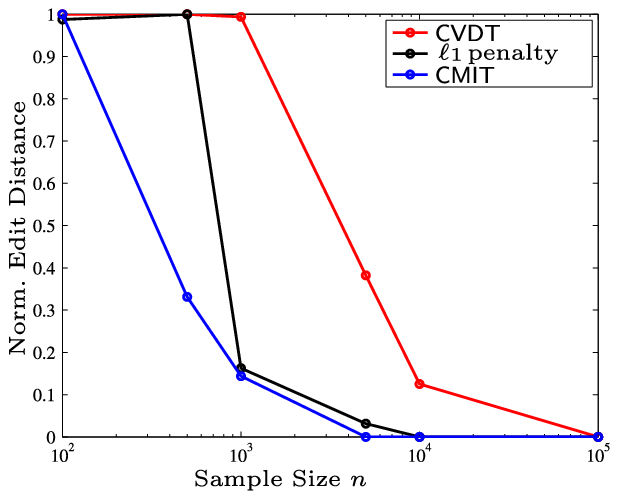}
}\\[-2pt]
\multicolumn{2}{c}{(c) Watts-Strogatz}
\end{tabular*}
%
\caption{$\threscondalgo$, $\threscmialgo$ and
$\ell_1$ penalized logistic regression on synthetic data from an
attractive Ising model.}\label{fig:attractive}
\end{figure}

\begin{figure}
 \begin{tabular*}{\textwidth}{@{\extracolsep{\fill}}cc@{}}

\includegraphics{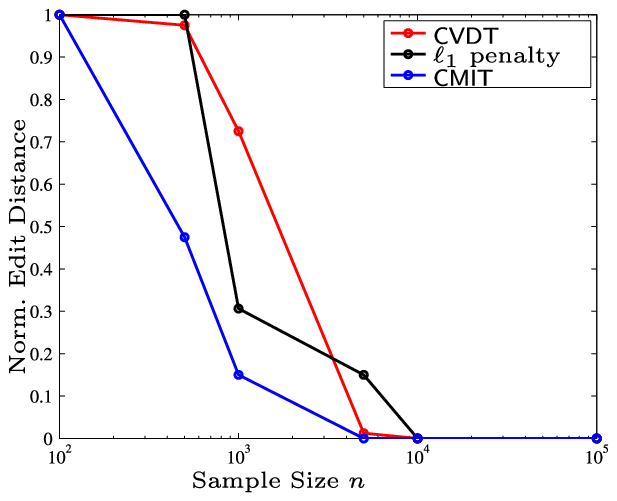}
&\includegraphics{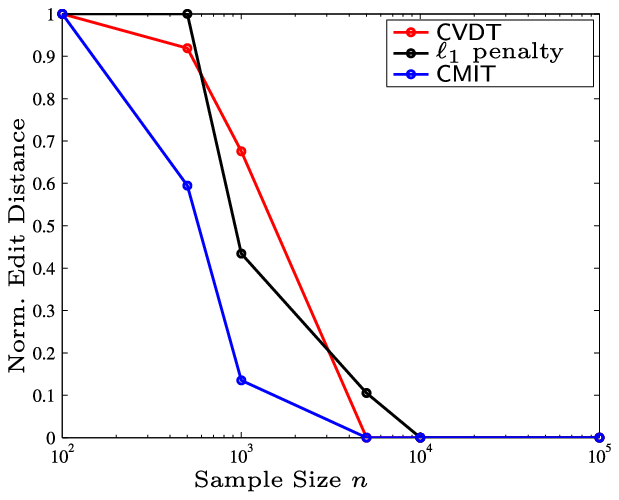}\\
(a) Cycle&(b) Erd\H{o}s-R\'{e}nyi
\\[6pt]
\multicolumn{2}{c}{
\includegraphics{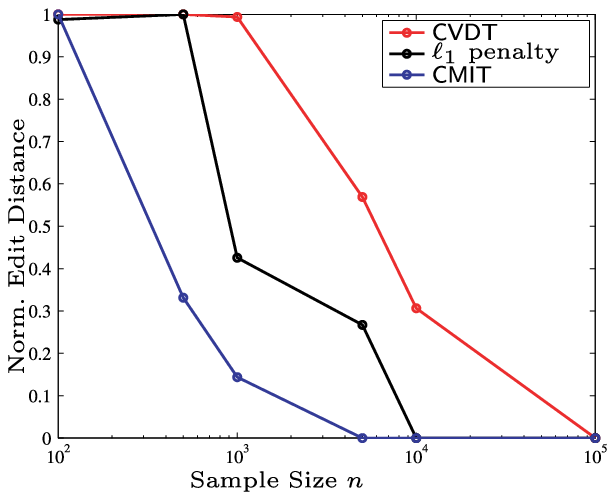}
}\\
\multicolumn{2}{c}{(c) Watts-Strogatz}
\end{tabular*}
%
\caption{$\threscondalgo$, $\threscmialgo$ and
$\ell_1$ penalized logistic regression on synthetic data from a mixed
Ising model (with both positive and negative edge
potentials).}\label{fig:nonattractive}
\end{figure}

Based on the generated graph topologies, we generate the potential
matrix $\mathbf{J }_G $ whose sparsity pattern corresponds to that of
the graph $G$. By convention, diagonal elements $\mathbf{J}(i,i)=0$ for
all $i \in V$. We consider both attractive and general models. For
attractive models, we consider the nonzero off-diagonal entries of
$\mathbf{J}$ as uniformly distributed in $[0.1,0.2]$. For the general
model, we consider the nonzero off-diagonal entries of $\mathbf{J}$ as
uniformly distributed in $[0.1,0.2]\cup[-0.1, -0.2]$. Potential
vector is set to $\mathbf{0}$ resulting in a symmetric Ising model.
Gibbs sampling method is used to generate samples. The knowledge of
the bound on local separators $\eta$ is assumed to be available in our
experiments. We employ normalized edit distances as the performance
criterion. Since we know the ground truth for synthetic data, it is
possible to evaluate this measure.
The thresholds $\xi_{n,p}$ for $\threscondalgo$/$\threscmialgo$ and
the regularization parameter $\lambda_n$ for the $\ell_1$ regularized
logistic regression are selected based on the best edit distances for
each method.

\subsection{Experimental results}

Table~\ref{table:collectionthreshold} presents the experimental
outcomes, and an explicit comparison of the three graph estimation
methods is illustrated in Figure~\ref{fig:attractive} for attractive
models, and in Figure~\ref{fig:nonattractive}
for mixed models (with both positive and negative edge potentials).
Similar trends are observed for both attractive and mixed models. We
note that the edit distance decays as the number of samples increases,
as expected. As long as there are enough number of samples (larger than
10,000), all the methods recover the graph structure accurately, that
is, with zero error. In terms of the decaying rate of errors, the $\ell
_1$ logistic regression method has a faster rate than $\threscondalgo$
for the Watts--Strogatz graph in all regimes, while for the cycle graph
and the Erd\H{o}s--R\'{e}nyi graph, the rates for $\threscondalgo$ and
the $\ell_1$ method are alternatively better depending on $n$.
However, $\threscmialgo$ has the fastest rate of decay of edit
distance for all the three graphs, although theoretically,
$\threscondalgo$ has better sample complexity guarantees compared to
$\threscmialgo$; see Theorem~\ref{thm:corr_thres_cond} and related
remarks. With regard to the running time, $\threscondalgo$ and
$\threscmialgo$ are faster for the graphs under consideration, since
there is one global threshold to be selected for finding all the edges,
while for logistic regression, selection of the regularization
parameter needs to be carried out for each neighborhood in the graph.
This is especially expensive for large graphs.

\section{Conclusion}

In this paper, we adopted a novel and a unified paradigm for Ising
model selection. We presented a simple local algorithm for structure
estimation with low computational and sample complexities under a set
of mild and transparent conditions. This algorithm succeeds on a wide
range of graph ensembles such as the Erd\H{o}s--R\'{e}nyi ensemble,
small-world networks etc. based on a local separation criterion.



\begin{supplement}[id=suppA]
\stitle{Supplement to ``High-dimensional structure estimation in Ising models: Local
separation criterion''}
\slink[doi]{10.1214/12-AOS1009SUPP} 
\sdatatype{.pdf}
\sfilename{aos1009\_supp.pdf}
\sdescription{Detailed analysis and proofs.}
\end{supplement}

\section*{Acknowledgments}
The authors thank Sujay Sanghavi (U.T.
Austin), Elchanan Mossel (UC Berkeley), Martin Wainwright (UC
Berkeley), Sebastien Roch (UCLA), Rui Wu (UIUC) and Divyanshu Vats (U.
Minn.) for extensive comments, and B\'ela Bollob\'as (Cambridge) for
discussions on random graphs. The authors thank the anonymous reviewers
and the co-editor Peter B\"uhlmann (ETH) for valuable comments that
significantly improved this manuscript.\vadjust{\goodbreak}

%

\printaddresses

\end{document}